\documentclass{article}

\usepackage[preprint]{neurips_2022}

\usepackage[utf8]{inputenc} 
\usepackage[T1]{fontenc}    
\usepackage{hyperref}       
\usepackage{url}            
\usepackage{booktabs}       
\usepackage{amsfonts}       
\usepackage{nicefrac}       
\usepackage{microtype}      
\usepackage{xcolor}         
\usepackage{graphicx}
\usepackage{subcaption}
\usepackage{caption}
\usepackage{amsmath}
\title{DiVAE : Photorealistic Images Synthesis with Denoising Diffusion Decoder}

\author{%
  Jie Shi\thanks{Both authors contributed equally to this research.}\\
  Peking University\\
  \texttt{jieshi@pku.edu.cn} \\
  \And
  Chenfei Wu$^{*}$\\
  Microsoft Research Asia \\
  \texttt{chewu@microsoft.com} \\
  \And 
  Jian Liang \\
  Peking University \\
  \texttt{j.liang@stu.pku.edu.cn} \\
  \And 
  Xiang Liu \\
  Peking University \\
  \texttt{xliu@ss.pku.edu.cn} \\
  \And
  Nan Duan\thanks{Corresponding author.} \\
  Microsoft Research Asia \\
  \texttt{nanduan@microsoft.com} \\
}

\begin{document}

\maketitle

\begin{abstract}
\vspace{-0.2cm}
  Recently most successful image synthesis models are multi stage process to combine the advantages of different methods, which always includes a VAE-like model for faithfully reconstructing embedding to image and a prior model to generate image embedding. At the same time, diffusion models have shown be capacity to generate high-quality synthetic images. Our work proposes a VQ-VAE architecture model with a diffusion decoder (DiVAE) to work as the reconstructing component in image synthesis. We explore how to input image embedding into diffusion model for excellent performance and find that simple modification on diffusion's UNet can achieve it. Training on ImageNet 256×256, Our model achieves state-of-the-art results and generates more photorealistic images specifically. In addition, we apply the DiVAE with an Auto-regressive generator on conditional synthesis tasks to perform  more human-feeling and detailed samples.
\vspace{-0.2cm}
\end{abstract}

\section{Introduction}
Recently, generative model of images, audio and videos have achieved rapid development and been capable of yielding impressive samples\cite{lao2019dual,li2019controllable,liang2020cpgan,zhang2021cross,tan2019semantics}. With promising future in application, the research of visual synthesis (images and videos) is becoming more and more popular.
\begin{figure*}[h]
    \vspace{-0.3cm}
    \centering
    \includegraphics[width=5.5in]{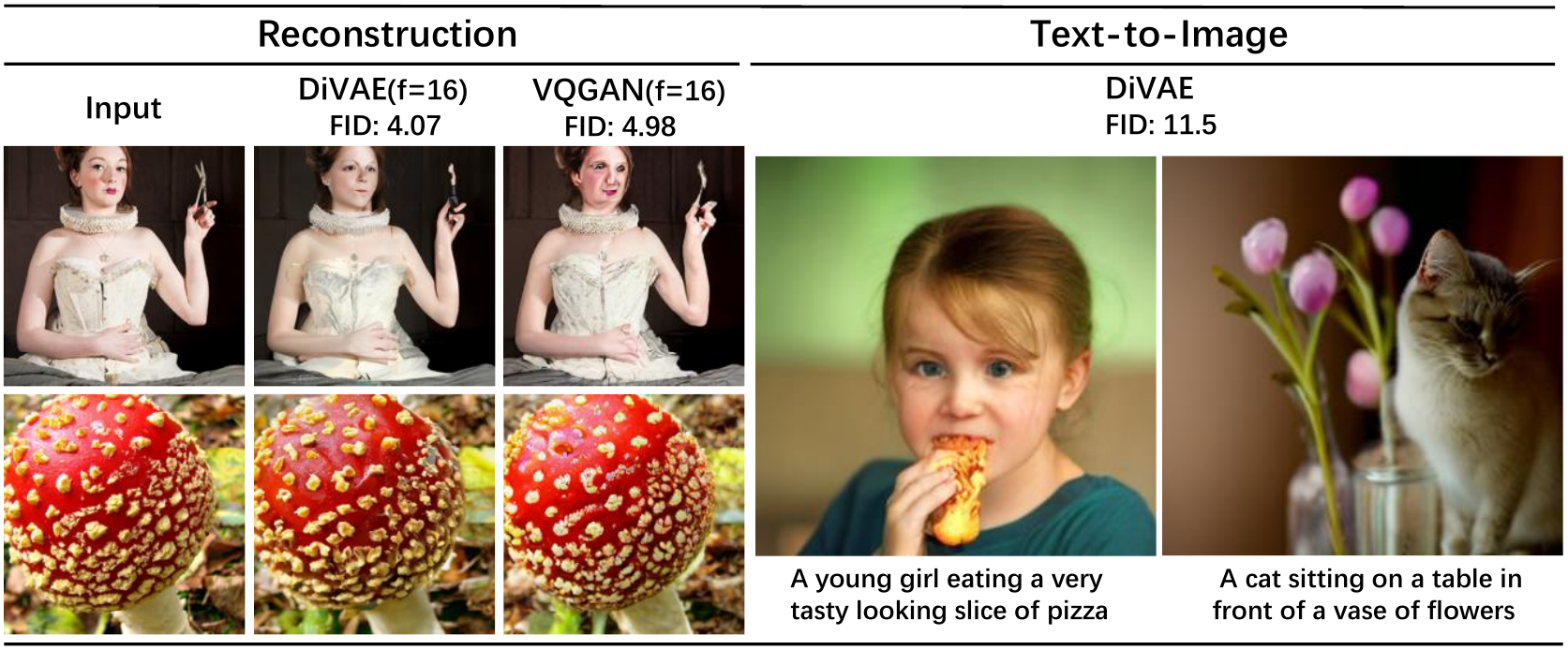}
    \caption{DiVAE generates more photorealistic and detailed images.}
    \label{fig1}
    \vspace{-0.2cm}
\end{figure*}

\textbf{Visual Synthesis} Several approaches have seen success in learning complex distributions of real vision, auto-regressive (AR) models\cite{brown2020language,chen2020generative,esser2021taming,salimans2017pixelcnn++,ding2021cogview,wu2021godiva,lin2021m6}, generative adversarial network(GANs)\cite{goodfellow2014generative,zhang2021cross,xu2018attngan,tao2020df,cha2019adversarial}, variational autoencoders (VAEs)\cite{kingma2013auto,rezende2014stochastic}, diffusion models\cite{sohl2015deep,ho2020denoising,nichol2021improved,nichol2021glide,rombach2021high,austin2021structured} and flow-based models\cite{dinh2014nice,chen2018neural,ho2019flow++,kingma2018glow} have shown convincing generative ability. Based on VAE, VQVQE\cite{van2017neural} and VQGAN\cite{esser2021taming} encode the image into a discrete latent space to learn the density of the hidden variables, and greatly improves the performance. In the past few years, GAN based model have shown their ability on high fidelity image generation and hold the state-of-the-art on many image generation tasks. However, GAN is often difficult to train and defective in capturing of diversity.Comparing with GAN, Auto-regressive (AR) generate model have advantages in density modeling and stable training. Based on the AR model, recent work like NUWA\cite{wu2021n}, DALL-E\cite{ramesh2021zero} and DALL-E2\cite{ramesh2022hierarchical}  have achieved impressive results on text-to-image, text-to-video generation, etc. 

\textbf{Diffusion models} are a class of likelihood-based models and have achieved state-of-the-art results in density estimation  as well as in sample quality\cite{ho2022cascaded}. The diffusion probabilistic models was introduced firstly in 2015\cite{sohl2015deep}, as a class of generative models which learn to match a data distribution by reversing a multi-step, gradual noising process. Denoising diffusion probabilistic models (DDPM)\cite{ho2020denoising} shows that diffusion models can produce high-quality images and the promising prospect in visual synthesis. After that, Improved DDPM\cite{nichol2021improved} 
 modified the learning of variance and optimization objectives to achieve better log-likelihoods. Denoising diffusion implicit models (DDIM)\cite{song2020denoising} developed a approach to fast sampling. Guided Diffusion\cite{dhariwal2021diffusion} find that samples from a class conditional diffusion model with a independent classifier guidance can be significantly improved. Classifier-Free Diffusion\cite{ho2021classifier} propose classifier-free guidance that does not need to train a separate classifier model.

\begin{figure*}[!t]
    \vspace{-0.5cm}
    \centering
    \includegraphics[width=5.3in]{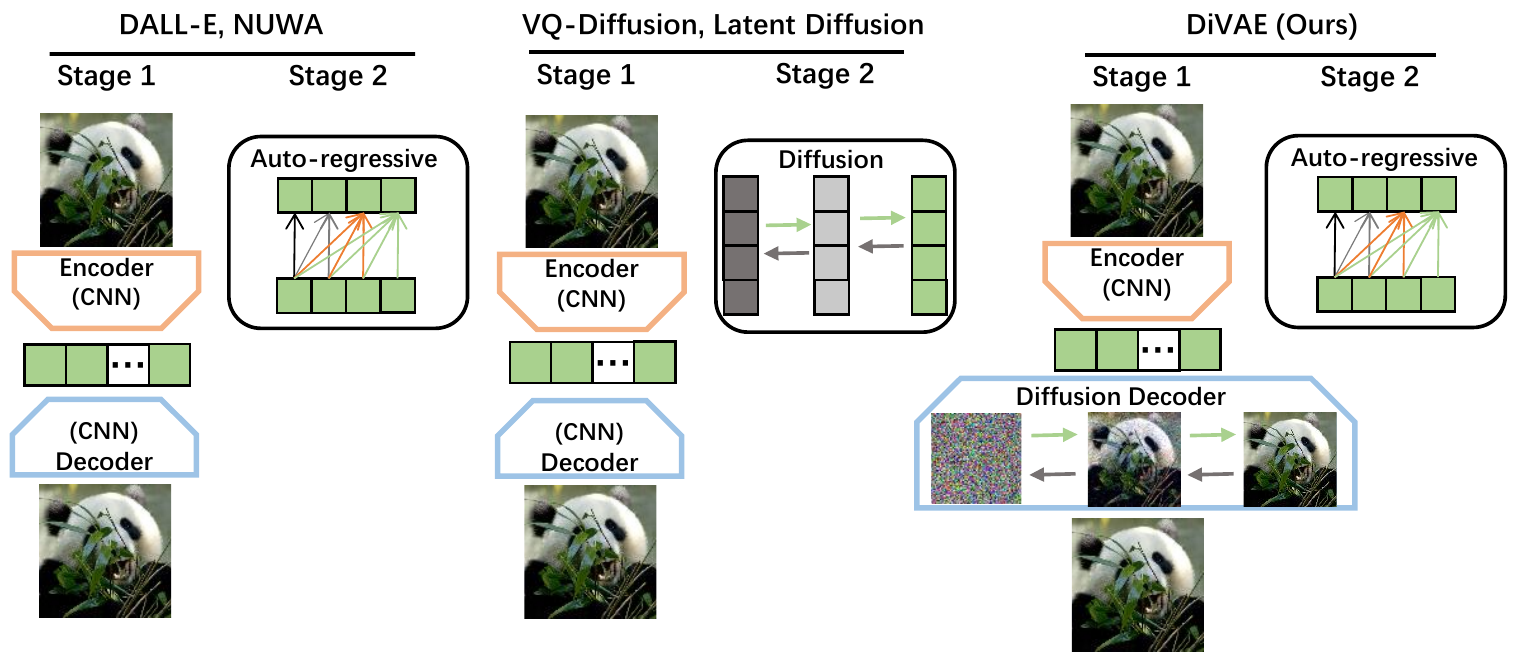}
    \caption{Overall framework of SOTA two stage image synthesis models, comparing the approach with DiVAE.}
    \label{fig2}
    \vspace{-0.2cm}
\end{figure*}

\textbf{One stage and Two stage Image Synthesis} Current visual generation work can be generalization into one-stage direct generation and two-stage generation\cite{dai2019diagnosing,rombach2020network,yu2021vector}. Visual auto-regressive models, such as PixelCNN\cite{van2016conditional}, PixelRNN\cite{van2016pixel}, and Image Transformer, diffusion model ,such as DDPM\cite{ho2020denoising} and Guided diffusion\cite{dhariwal2021diffusion}, performed visual synthesis in a “pixel-by-pixel” manner, optimization and inference often is with high computational cost. To mitigate the shortcomings of individual approaches, a lot of research combine the strengths of different methods. As Figure \ref{fig2} shows, DALL-E\cite{ramesh2021zero}, NUWA\cite{wu2021n}, VQ-Diffusion\cite{gu2021vector} and Latent Didffusion\cite{rombach2021high} first learn an encoder-decoder architecture, like VQ-GAN\cite{esser2021taming} and VQ-VAE\cite{van2017neural}, which can compress image to latent representation and faithfully reconstruct it back to image, in second stage, AR based model: DALL-E\cite{ramesh2021zero} and NUWA\cite{wu2021n} sequentially predict image token depends on the condition, while diffusion based model, VQ-Diffusion\cite{gu2021vector} and Latent Didffusion\cite{rombach2021high} predict it with a gradual denoising process.

\textbf{Our work} aims to propose an generality model to generate more detailed and photorealistic images to improve the reconstructing stage of multi stage image synthesis. The potential improvements of image embedding generating model is expecting for future works.

$\bullet$ We propose DiVAE, a vae generation framework with a diffusion decoder, which can generate more photorealistic images and achieved state-of-the-art results on image reconstruction from embeddings.
$\bullet$ We perform DiVAE with an Auto-regressive generator on Text-to-Image (T2I) synthesis tasks and generate high-quanlity and more detailed images. 

\section{Background}
\label{Background}

\subsection{Diffusion Models' Details}

Diffusion generative models were first introduced by in 2015\cite{sohl2015deep} and improved in denoising diffusion probabilistic models (DDPM)\cite{ho2020denoising} and Improved Denoising Diffusion Probabilistic Models\cite{nichol2021improved}, achieved state-of-the-art results on image generation. It has recently been shown that this class of models can produce high-quality images and 
have been researched in a series of recent visual\cite{ramesh2022hierarchical,nichol2021glide,rombach2021high} and audio\cite{chen2020wavegrad,kong2020diffwave} synthesis tasks.

In diffusion model, the diffusion process (forward process) starts from data distribution ${x}_{0} \sim q({x}_{0})$ and a Markovian noising process $q$ which gradually adds noise to the data to produce noised samples ${x}_{1}$ through ${x}_{T}$, each step of the noising process adds Gaussian noise according to variance schedule given by ${\beta}_{t}$, until ${x}_{T}$ is nearly an isotropic Gaussian distribution. We define ${\alpha}_{t} = 1- {\beta}_{t}$ and ${\hat{\alpha}}_{t} = \prod_{s=0}^{t} \alpha_{s}$, the diffusion process can be expressed as:

\begin{equation}
q({x}_{t}|{x}_{t-1}) \sim \mathcal N ({x}_{t};\sqrt{1-{\beta}_{t}}{x}_{t-1};{\beta}_{t}I)
\end{equation}

\begin{equation}
q({x}_{t}|{x}_{0}) \sim \mathcal N ({x}_{t};\sqrt{{\hat{\alpha}}_{t}}{x}_{0};(1-{\hat{\alpha}}_{t})I) = \sqrt{{\hat{\alpha}}_{t}}{x}_{0} + \sqrt{\sigma 1-{\hat{\alpha}}_{t}}, \sigma \sim \mathcal N (0,I)
\end{equation}

\begin{equation}
{x}_{t} = \sqrt{{\hat{\alpha}}_{t}}{x}_{0} + \sqrt{\sigma 1-{\hat{\alpha}}_{t}}, \sigma \sim \mathcal N (0,I)
\end{equation}

$1-{\alpha}_{t}$is variance of the noise for an arbitrary timestep, and we could equivalently use this to define the noise schedule instead of ${\beta}_{t}$.With Bayes theorem,  the posterior $q({x}_{t-1}|{x}_{t}, {x}_{0})$ in terms of ${\beta}_{t}$  and $\hat{\mu} ({x}_{t}, {x}_{0})$ can be expressed as:

\begin{equation}
q({x}_{t-1}|{x}_{t}, {x}_{0}) \sim \mathcal N ({x}_{t-1};\hat{\mu} ({x}_{t}, {x}_{0});{\hat{\beta}}_{t}I)
\end{equation}

\begin{equation}
{\hat{\beta}}_{t} = \frac{1-{\hat{\alpha}}_{t-1}}{1-{\hat{\alpha}}_{t}}{\beta}_{t}
\end{equation}
\begin{equation}
\mu ({x}_{t}, {x}_{0}) = \frac{\sqrt{{\hat{\alpha}}_{t-1}}{\beta}_{t}}{1-{\hat{\alpha}}_{t}}{x}_{0} + \frac{\sqrt{{\alpha}_{t}}(1-{\hat{\alpha}}_{t-1})}{1-{\hat{\alpha}}_{t}}{x}_{t}
\end{equation}

It can be seen from the above, when the exact reverse distribution $q({x}_{t-1}|{x}_{t})$ is known, ${x}_{T}$ can be sampled from $\mathcal N (0,I)$, and then we can get a sample from $q({x}_{0})$ from running the reverse process.In particular, sampling starts with noise ${X}_{T}$ and produces gradually less-noisy samples ${x}_{T-1}, {x}_{T-2}, ... $ until reaching a final sample ${X}_{0}$ . However, since $q({x}_{t-1}|{x}_{t})$ depends on the entire data distribution, diffusion model approximate it using a neural network as follows ${p}_{\theta}({x}_{t-1}|{x}_{t})$. A diffusion model learns to produce a slightly more denoised ${X}_{t-1}$from ${X}_{t}$. 

\begin{equation}
{p}_{\theta}({x}_{t-1}|{x}_{t}) \sim \mathcal N ({x}_{t-1};\mu ({x}_{t},t), \sum({x}_{t},t))
\end{equation}

The most obvious option is to predict ${\mu}_{\theta} ({x}_{t},t)$  directly with a neural network. Alternatively, the network could predict ${X}_{0}$, and this output could be used in Equation
5 to produce ${\mu}_{\theta} ({x}_{t},t)$. DDPM found that a different objective produces better samples in practice. In particular, they do not directly parameterize  ${\mu}_{\theta} ({x}_{t},t)$ as a neural network, but instead train a model $\theta({x}_{t},t)$ to predict.

\begin{equation}
\mu ({x}_{t}, {x}_{0}) = \frac{1}{\sqrt{{\alpha}_{t}}} ({x}_{t} - \frac{1-{\alpha}_{t}}{\sqrt{1-{\hat{\alpha}}_{t}}} {\epsilon}_{t}({x}_{t}, {x}_{0}))
\end{equation}

In obvious works, $\sum_{\theta} ({x}_{t},t)$ often is  constraints not learned, DDPM finds that its reasonable range is very small, and it would be hard for a neural network to predict $\sum ({x}_{t},t)$ directly. Improved Denoising Diffusion Probabilistic Models proposes to parameterize the variance as an interpolation between ${\beta}_{t}$ and ${\hat{\beta}}_{t}$ in the log domain. In this work, model outputs a vector $v$ containing one component per dimension, and we turn this output into variances as follows:

\begin{equation}
\sum_{\theta} ({x}_{t},t) = exp(vlog{\beta}_{t} + (1-v)log{\hat{\beta}}_{t})
\end{equation}

\subsection{VQ-VAE Architectures Models}
Auto-regressive transformer architectures have shown great promise in image synthesis due to their outstanding expressivity. Since the computation cost is quadratic to the sequence length, it is computationally prohibitive to directly model raw pixels. To reduce the description length of compositions, recent works propose to represent an image by discrete image tokens with reduced sequence length. Hereafter a transformer generator can be effectively trained upon this reduced context length or image tokens. With a encoder-decoder architectures and a codebook, the image can be compressed to latent representation via the CNN encoder and then faithfully reconstructed via the CNN decoder.Vector quantized variational autoencoder (VQ-VAE)\cite{van2017neural} consists of an encoder $E$, a decoder $G$ and a codebook $Z = {{z}_{k}}_k=1^K \in {R}^K×d$containing a finite number of embedding vectors ,where $K$ is the size of the codebook and $d$ is the dimension of codes. Given an image $x \in {R}^{HW3},$ VQ-VAE obtain a spatial collection of image tokens ${z}_{q}$ with the encoder $z = E(x)  \in {R}^{hwd}$and a subsequent spatial-wise quantizer Q() which maps each spatial feature  ${z}_{ij}$into its closest codebook entry ${z}_{k}$:

\begin{equation}
{z}_{q} = Q(z) = (\mathop{\arg\min}\limits_{{z}_{k} \in Z} ||{\hat{z}}_{ij}-{z}_{k}||) \in {R}^{h×w×{n}_{z}}
\end{equation}
The reconstructing is performed by a CNN decoder, and reconstruction $\hat{x} = x$ is
\begin{equation}
\hat{x} = G({z}_{q}) = G(q(E(x)))
\end{equation}
Backpropagation through the non-differentiable quantization operation is achieved by a straight-through gradient estimator(STE) to train encoder and decoder, as well as learning an effective codebook.
\begin{equation}
{L}_{VQ}(E,G,Z) = {||x-\hat{x}||}^{2} + {||sg[E(x)]-{z}_{q}||}_2^2 + \beta{||sg[{z}_{q}]-E(x)||}_2^2
\end{equation}

In order to learn a richer codebook. VQ-GAN\cite{esser2021taming},a variant of the original VQ-VAE, and use a discriminator and perceptual loss to keep good
perceptual quality at increased compression rate, was proposed. An adversarial training with a patch-based discriminator D that aims to differentiate between real and reconstructed images was used:
\begin{equation}
{L}_{GAN}({E,G,Z},D) = [logD(x)+log(1-D(\hat{x}))]
\end{equation}
And the complete object of VQGAN is:
\begin{equation}
[{L}_{VQ}(E,G,Z)+\lambda {L}_{GAN}({E,G,Z},D)]
\end{equation}
\section{DiVAE}
\label{method}
\subsection{Denoising Diffusion Decoder in DiVAE}
Perhaps the work most related to our approach are VQ-VAE\cite{van2017neural} and VQ-GAN\cite{esser2021taming}, which consist of the following parts: a CNN encoder which parameterizes a posterior distribution $q(z|x)$ of discrete latent variables $z$ from input data $x$, a Codebook $Z$ containing a finite number of embedding vectors, and a decoder with a distribution $p(x|z)$, and the image can be faithfully reconstructed via the CNN decoder. In multi-stage image synthesis, the reconstructing decoder greatly influence the quality of results.

\begin{figure*}[!t]
    \centering
    \includegraphics[width=5.5in]{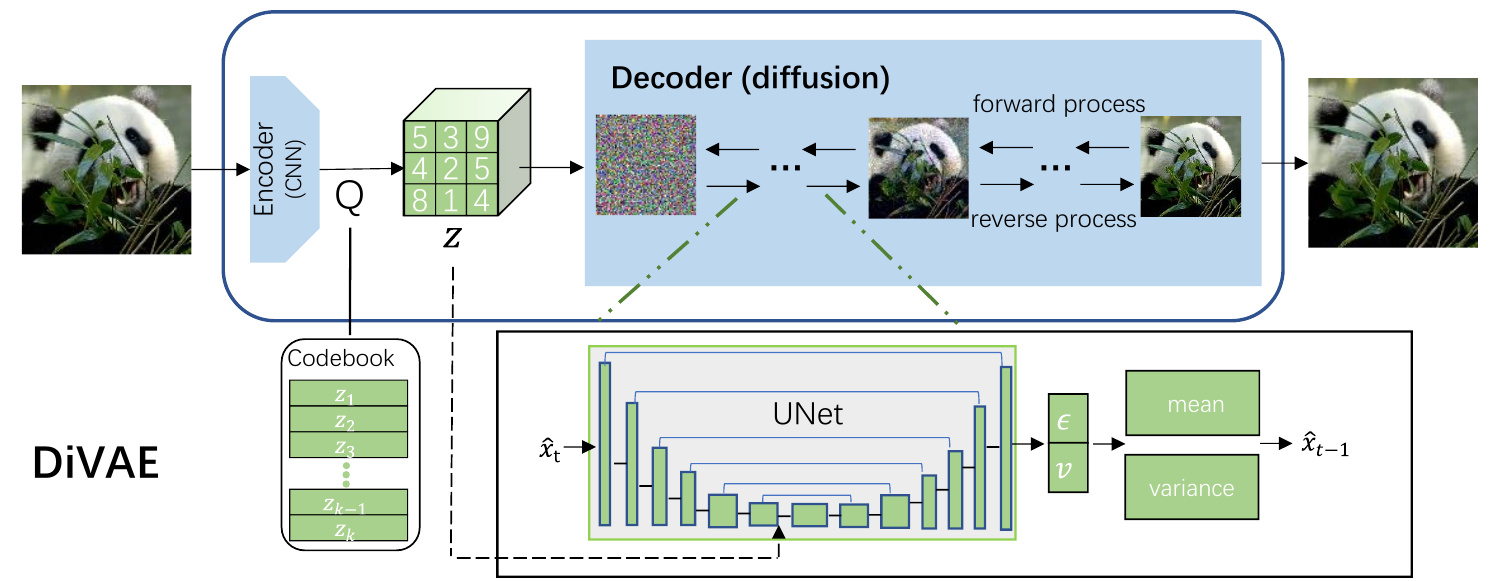}
    \caption{Structure of DiVAE, which uses a convolutional encoder and a denoising diffusion decoder, and the key Network is an UNet to model $p({x}_{t-1}|{x}_{t},z)$ }
    \label{fig3}
\end{figure*}

Different from VQ-GAN and VQ-VAE, DiVAE parameterizes $p(x|z)$ through a denoising diffusion model, take DDPM's denoising process for instance, $p(x|z)$ can be obtained by $T$ times’ iterative $p({x}_{t-1}|{x}_{t},z)$, in which sampling starts with noise ${x}_{T}$ and produces gradually less-noisy samples ${x}_{t-1}{x}_{t-2}$ ... until reaching a final sample ${x}_{0}$:

\begin{equation}
p(x|z) = p({x}_{0}|z) = \prod_{t=T}^{1} p({x}_{t-1}|{x}_{t},z)
\end{equation}
DiVAE still is a VQ-VAE architecture model, a diffusion decoder generates images conditiond on images embeddings and a CNN encoder can compress images to embeddings. We modify the architecture proposed in Improved DDPM\cite{nichol2021improved} and some UNet improvement in Guided Diffudion\cite{dhariwal2021diffusion}, inputs 16×16 or 32×32 image embeddings into the middle block of UNet. Diffusion model can generate high-quality image via GAN, thus we hope that our model can reconstruct high-quality image from latent representation $z$ with a denoising diffusion decoder, and result in Section 4 shows that DiVAE achieves goals. 

In training process, DiVAE's diffusion process is same as diffusion model's. The forward process or diffusion process is a noising process $q({x}_{t}|{x}_{t-1})$ which is fixed to a Markov chain that gradually adds Gaussian noise to the data according to a variance schedule ${\beta}_{1}, . . . , {\beta}_{T}$. According to Bayes theorem, one finds that the reverse process , the posterior $q({x}_{t-1}|{x}_{t}, {x}_{0})$ is also a Gaussian with mean ${\hat{\mu}}_{t}({x}_{t},{x}_{0})$ and variance ${\hat{\mu}}_{t}$ . 

We aim to model the denoise diffusion process $p({x}_{t-1}|{x}_{t}, z)$ to estimate the posterior transition distribution $q({x}_{t-1}|{x}_{t}, {x}_{0})$. From previous diffusion model, the key Network is an U-Net architecture that is trained with a denoising objective to iteratively remove various levels of noise from the input. The current timestep $t$ is injected into the network with Adaptive Group Normalization (AdaGN) operator, i.e., $AdaGN(h, t) = {a}_{t}LayerNorm(h) + {b}_{t}$, where $h$ is the intermediate activations, at and ${b}_{t}$ are obtained from a linear projection of the timestep embedding.

\begin{equation}
{p}_{\theta}({x}_{t-1}|{x}_{t}, z) \sim \mathcal N ({x}_{t-1};\mu ({x}_{t},t), \sum ({x}_{t},t))
\end{equation}

Essentially the work is to parameterize the model as a function to predict ${\mu}_{\theta} ({x}_{t},t)$ which define the noise component of a noisy sample ${x}_{t}$ and the $\sum_{\theta} ({x}_{t},t)$ define the variance. DDPM observe that the simple mean-sqaured error objective, ${L}_{simple}$, which can be seen as a reweighted form of ${L}_{vlb}$, works better in practice than the actual variational lower bound ${L}_{vlb}$.  

\begin{equation}
{L}_{simple} = {E}_{t\sim [1,T],{x}_{0} \sim q({x}_{0}),\epsilon \sim \mathcal N (0,I)} [{||\epsilon - {\epsilon}_{t}({x}_{t}, {x}_{0})||}^{2}]
\end{equation}

\begin{equation}
{L}_{vlb} = {L}_{0} + {L}_{1} + {L}_{2} + ... + {L}_{T-1} + {L}_{T}
\end{equation}
\begin{equation}
{L}_{0} = -log{p}_{\theta}({x}_{0}|{x}_{1},y)
\end{equation}
\begin{equation}
{L}_{t-1} = {D}_{KL}(q({x}_{t-1}|{x}_{t},{x}_{0})||{p}_{\theta}({x}_{t-1}|{x}_{t},z))
\end{equation}
\begin{equation}
{L}_{T} = {D}_{KL}(q({x}_{T}|{x}_{0})||p({x}_{T})
\end{equation}

Our work follows the hybrid objective in Improved DDPM, as early DDPM set $\sum({x}_{t},t)$ not learned, Improved DDPM experiments and considers two opposite extremes, parameterized the variance as an interpolation and achieved their best results. Since ${L}_{simple}$ doesn’t depend on $\sum ({x}_{t},t)$, they use an new hybrid objective:
\begin{equation}
{L}_{hybrid} = {L}_{simple} + \lambda {L}_{vlb} 
\end{equation}

\subsection{Application in Conditional Images Synthesis Tasks}

Previous multi stage image synthesis models e.g. DALL-E, NUWA, includes a VAE structure model for faithfully reconstructing image embedding to image and an auto-regressive model sequentially predict image tokens to generate image embedding. Our work prevents such two stage visual synthesis trade-offs, improves the first stage ability by realizing a diffusion decoder based on Denoising Diffusion Model. With a denoising diffusion decoder, DiVAE can reconstruct more high-quality and detailed image from a latent representation $z$. We apply the DiVAE with an auto-regressive generator on Text-to-Image tasks to evaluate our model's ability and generality of better and more human-feeling samples. The auto-regressive generator is an transformer encoder-decoder framework covering language and image to realize conditioned synthesis. The model's target is to model p(x|y), which can be mianly divided into the modeling and optimization of token generation and image reconstruction:

\begin{equation}
\begin{aligned}
p(x|y):\begin{cases}
p(z|y) = \prod_{i=1}^{N} p({z}^{i}|{z}^{1},...,{z}^{i-1},y) \\
p(x|z) = \prod_{t=T}^{1} p({x}_{t-1}|{x}_{t},z)
\end{cases}
\end{aligned}
\end{equation}

Auto-regressive generator produce $z$ from captions to enable image embedding generations from text captions, and hen diffution decoder of DiVAE reconstruct it to synthesis image. 

\section{Experiments}
\label{exp}
This section evaluates the ability of our approach to reconstruct high-quality image from image embedding, we use Fréchet Inception Distance (FID)\cite{heusel2017gans} as our default metric as the other state-of-the-art generative modeling works. Firstly, We compare the DiVAE's performance of reconstructing with the state-of-the-art generative models in Sec \ref{sec4_1}. In addition, we apply the DiVAE with a pre-trained Auto-regressive generator on Text-to-Image (T2I) synthesis task to evaluate its generality in visual synthesis tasks in Sec \ref{sec4_2}. Ablation Study in Sec \ref{sec4_3} explores how to add image embedding into diffusion model for excellent performance. We train our model on ImageNet datasets\cite{deng2009imagenet} with a training batch size of 256 for total 10k training steps, AdamW optimizer is used with the learning rate linearly warming up to a peak value of 1e-4 over 1000 steps.

\subsection{Comparison with state-of-the-art}
\label{sec4_1}

\begin{table*}[h]
\setlength{\tabcolsep}{0.16in}
\label{table3}
\begin{center}
\begin{tabular}{lcccccr}
\toprule
Model & Dataset & $D \rightarrow R$ & Rate & dim $Z$ &FID \\
\midrule
DALL-E dVAE & Web data & $32^2 \rightarrow 256^2$ & f8 & 8192 &32.0 \\
VQGAN & ImageNet & $16^2 \rightarrow 256^2$ & f16 & 1024&7.94 \\
VQGAN & ImageNet & $16^2 \rightarrow 256^2$ & f16 & 16384&4.98 \\
VQGAN & ImageNet & $32^2 \rightarrow 256^2$ & f8 & 8192&1.49\\
\midrule
DiVAE(ours) & ImageNet & $16^2 \rightarrow 256^2$ & f16 & 16384& \textbf{4.07}\\
DiVAE(ours) & ImageNet & $32^2 \rightarrow 256^2$ & f8 & 8192& \textbf{1.24} \\
\bottomrule
\end{tabular}
\end{center}
\vskip -0.05in
\caption{FID between reconstructed validation split and original validation with 50000 images split on ImageNet.}
\label{restruct}
\end{table*}

\begin{figure*}[t]
    \centering
    \includegraphics[width=5.5in]{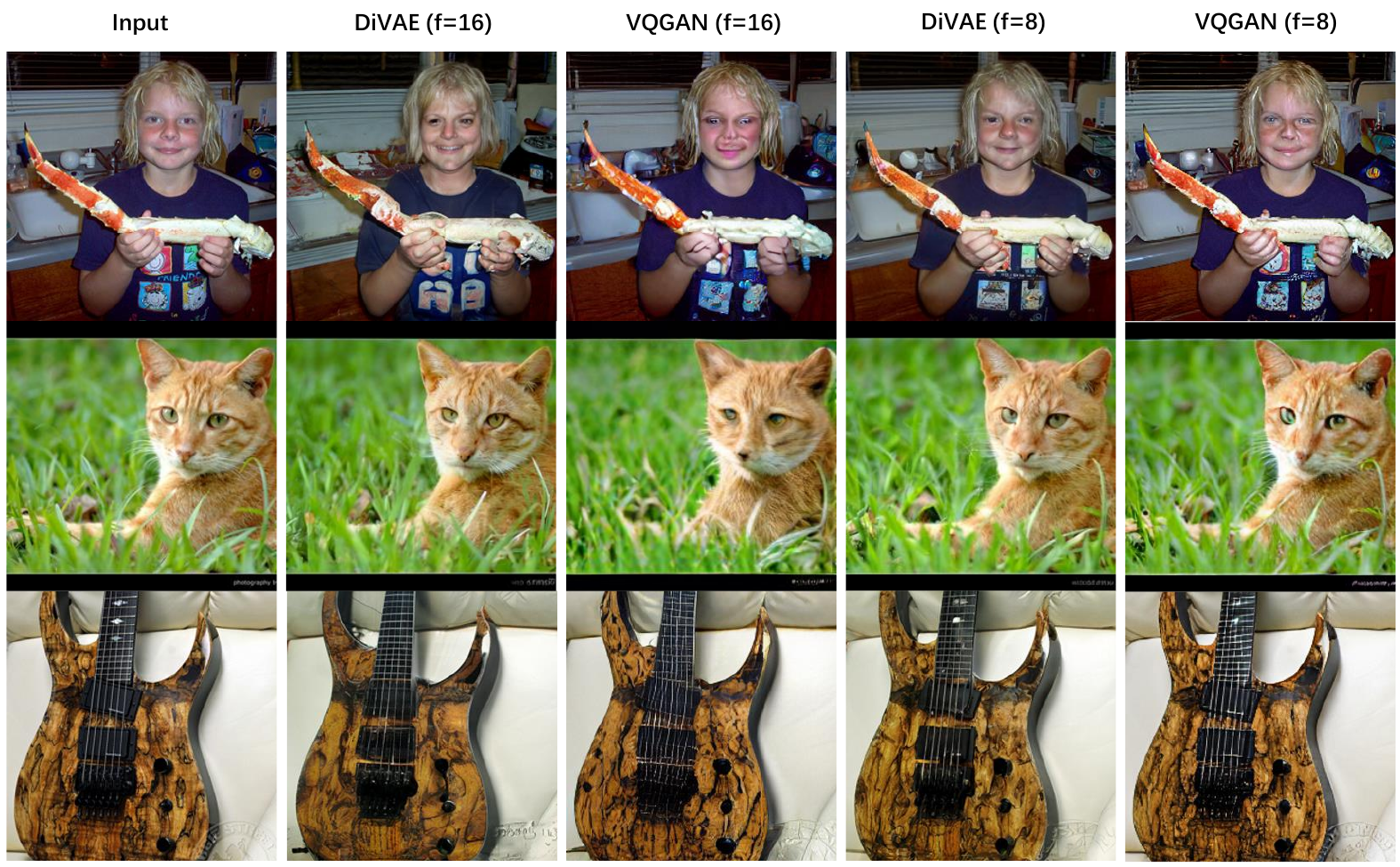}
    \caption{Comparison with VQGAN in $32^2 \rightarrow 256^2$ and $16^2 \rightarrow 256^2$ reconstruction with same codebook dimension $Z$.}
    \label{fig4}
\end{figure*}

We investigate how our approach quantitatively and qualitatively compares to existing state-of-the-art models (VQVAE,VQGAN) for generative image synthesis. The model are trained in two compression rate (f16 and f8). In our training, model built with 1000 diffusion steps, in inference, DiVAE model use 1000 steps DDPM sampling approaches. Table \ref{restruct}shows FID between reconstructed images and original images in the validation split on ImageNet, DiVAE is able to achieve better FID compared with VQGAN in case of both $32^2 \rightarrow 256^2$ and $16^2 \rightarrow 256^2$ generating, achieving state-of-the-art results. The diffusion model bring its ability of high-quality sampling to DiVAE. Specifically, as show as Figure \ref{fig4}, improvement is more obvious in  human-feeling quality and details of images. DiVAE's synthesis sample is more photo-realistic and reconstructing well more details, like eyes and small target, the improvement is significantly especially in Rate of f16.

\subsection{Text-to-Image (T2I) Task}
\label{sec4_2}

Our work aims to propose an generality model to generate more realistic and photo-realistic images to improve the reconstructing stage of multi stage image synthesis, so the first stage of text-to-image model in our experiment still is an auto-regressive transformer to generate image embedding conditioned on text. We pre-train an auto-regressive transformer on Conceptual Captions\cite{lin2014microsoft} for text-to-image (T2I) generation. The  auto-regressive model produces image embedding from text, and the $32^2 \rightarrow 256^2$ diffusion decoder reconstructs image conditioned on $z$ inverting the compressing process of encoder.

\begin{figure*}[h]
    \centering    \includegraphics[width=5.5in]{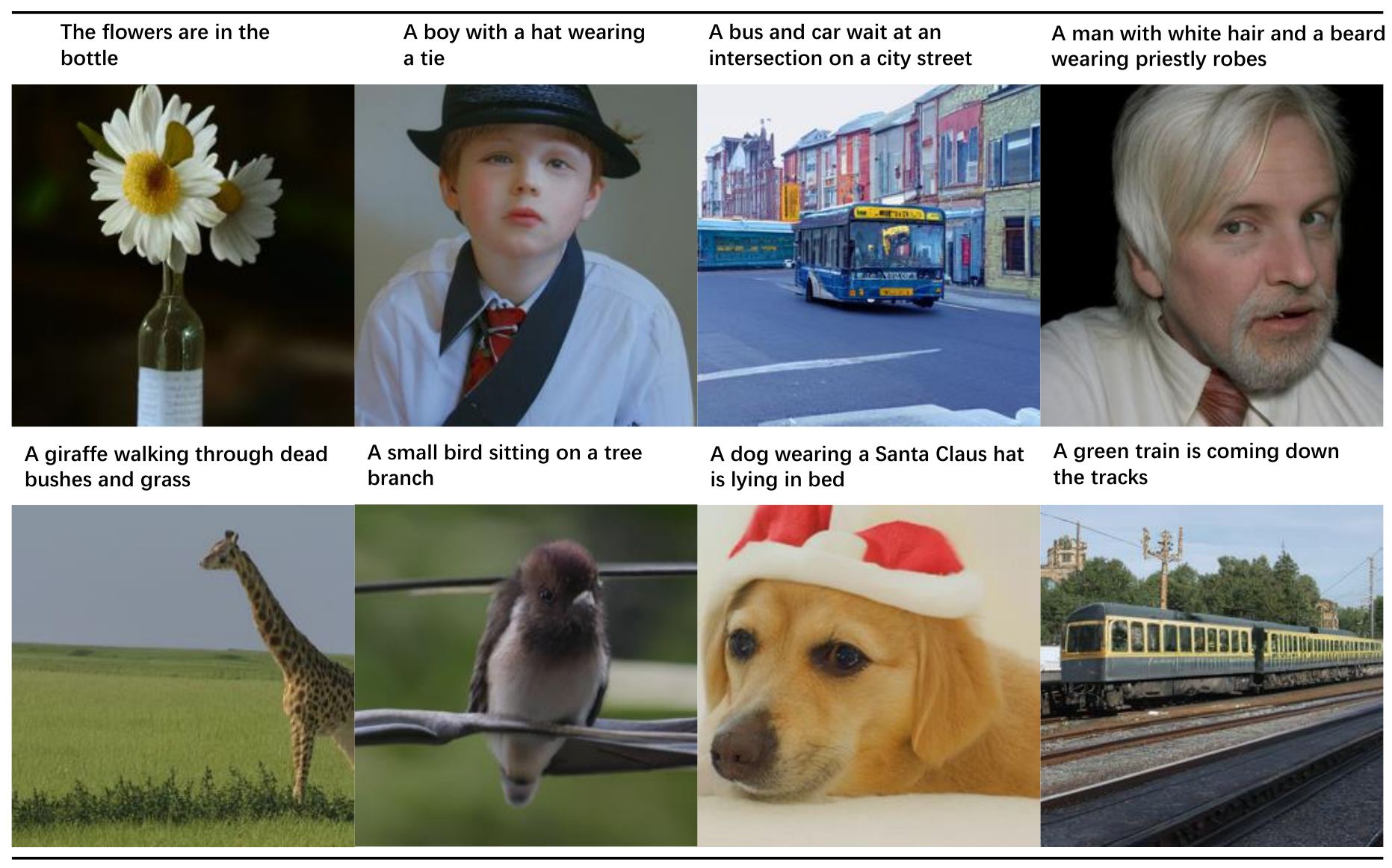}
    \caption{Samples of Text-to-Image (T2I) task generated by DiVAE.}
    \label{fig5}
\end{figure*}

\begin{table*}[h]
\setlength{\tabcolsep}{0.41in}
\begin{center}
\begin{tabular}{lccr}
\toprule
Model &   FID & Zero-shot FID \\
\midrule
AttnGAN\cite{xu2018attngan} & 35.49  & \\
DM-GAN\cite{zhu2019dm} &  32.64 & \\
DF-GAN\cite{tao2020df} &  21.42 & \\
DALL-E\cite{ramesh2021zero} &  27.50 & \\
CogView\cite{ding2021cogview} & 27.10  & \\
NUWA\cite{wu2021n} & 12.9  & 22.6 \\
VQ-Diffusion\cite{gu2021vector} & 13.86 & \\
XMC-GAN\cite{zhang2021cross} & 9.33 &  \\
GLIDE\cite{nichol2021glide} &   & 12.24 \\
DALL-E2\cite{ramesh2022hierarchical} &   & 10.39 \\
\midrule
DIVAE (Ours) & 11.53 & \\
\bottomrule
\end{tabular}
\end{center}
\vskip -0.05in
\caption{Qualitative comparison with the state-of-the-art models for Text-to-Image (T2I) task on the MSCOCO (256×256) dataset.}
\label{t2i}
\end{table*}

\begin{figure*}[h]
    \centering    \includegraphics[width=5.5in]{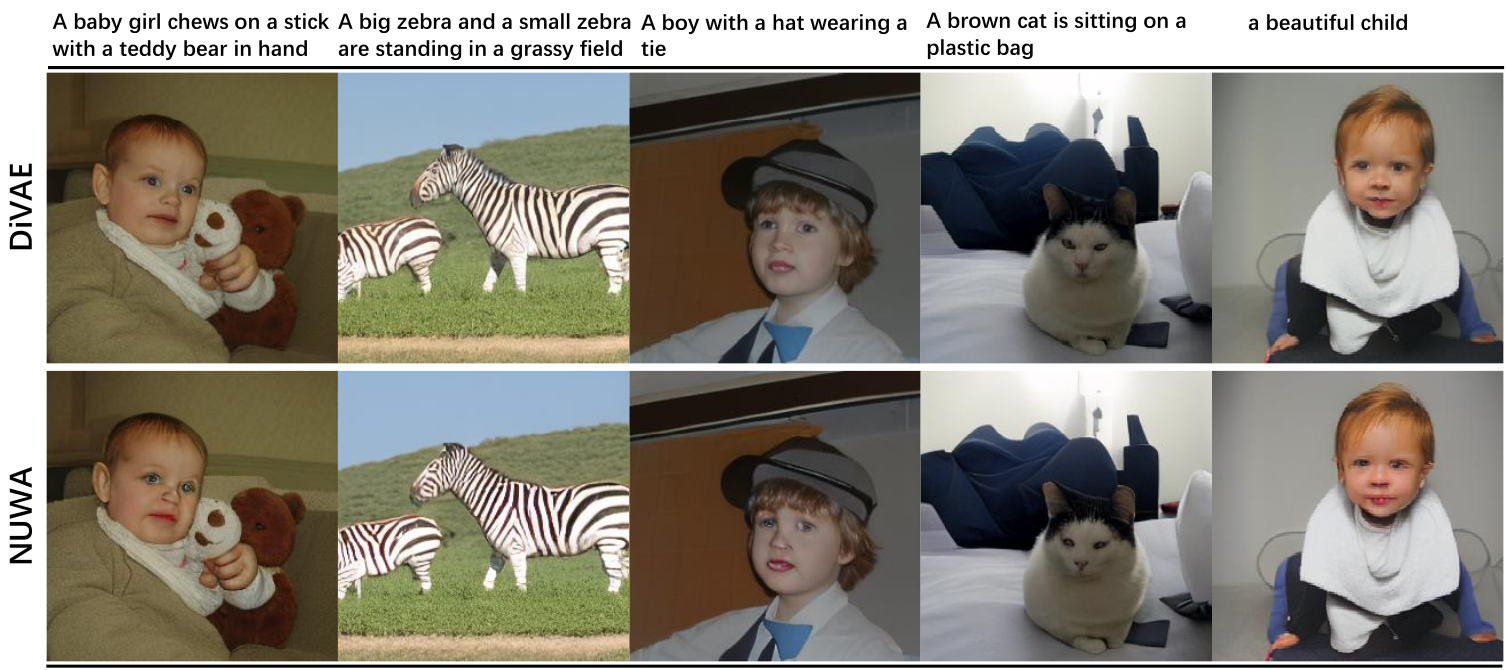}
    \caption{Qualitative comparison NUWA for Text-to-Image (T2I) task, with the same image embeddings from NUWA.}
    \label{fig6}
\end{figure*}

Comparing in Tabel \ref{t2i}, we perform text-to-image prediction quantitatively on MSCOCO (256×256) datasets\cite{lin2014microsoft} and qualitatively show results in Figure \ref{fig5}. As show in Tabel \ref{t2i},FID of GLIDE and DALLE-2 is really perfect, GLIDE is an one-stage classifier-free guidance diffusion model, DALL-E2 is an muti stage visual synthesis model with an excellent prior and an classifier-free guidance diffusion\cite{ho2021classifier} decoder. Both of them have an better generator from text and huge high-quality training datasets. Except them, DiVAE achieves FID 11.53, is better than previous models. Figure shows the synthesized images of DiVAE conditioned on text in MSCOCO validation, obviously, the samples are realistic and have fine-grained details. 

NUWA is an two stage visual synthesis model, which also contains an auto-regressive generator and an embedding decoder.As comparison in NUWA, although reporting a significant FID score of 9.3, XMC-GAN's sampling images are not realistic like NUWA, so we compare our synthesis images with NUWA in same condition in Figure \ref{fig6}. In the experiment, we take the $16^2$ image embeddings from NUWA's auto-regressive model to DiVAE's decoder to reconstruct $256^2$ size for clearer comparison in Figure . DiVAE's synthesis images is more photorealistic and reconstructing well more details, as the eyes and hail are significantly fine-reconstructed. The phenomenon will be more obvious in $16^2 \rightarrow 256^2$ decoder's comparison as show in Table \ref{restruct} and Figure \ref{fig4}, The ability of DiVAE allows larger compressing ratio of image in  multi stage visual synthesis model, leading to more efficient and large-scale pre-training.

\subsection{Ablation Study}
\label{sec4_3}

The key of DiVAE is how to add image embeddings to diffusion model's Network, our work explores several approaches to make it work better. In this section we comparing the FID of different setting on 5000 validation of ImageNet. UNet for diffusion modeling is an encoder-decoder Network, so we investigate to concat embeddings to the encoding blocks, middle blocks and decoding blocks to train the model. As the results in Table \ref{tab3_2}, the FID score of adding to middle block is better than the others, considering the middle block of UNet\cite{ronneberger2015u} is the bottom of features with more concentrated information.

\begin{minipage}[!t]{0.5\textwidth}
\centering
\setlength\tabcolsep{1.3mm}
\begin{tabular}{lccc}
\toprule
method & concat & add & attention  \\
\midrule
FID & 11.58 & 11.61 & 13.35 \\
\bottomrule
\end{tabular}
\captionof{table}{Method of inputting embeddings.}
\label{tab3_1}
\end{minipage}
\begin{minipage}[!t]{0.5\textwidth}
\centering
\setlength\tabcolsep{1mm}
\begin{tabular}{lccc}
\toprule
position & encoder & middle & decoder   \\
\midrule
FID & 13.08 & 11.58 & 13.06 \\
\bottomrule
\end{tabular}
\captionof{table}{Position of inputting embeddings.}
\label{tab3_2}
\end{minipage}

\begin{figure*}[h]
    \centering
    \includegraphics[width=5.5in]{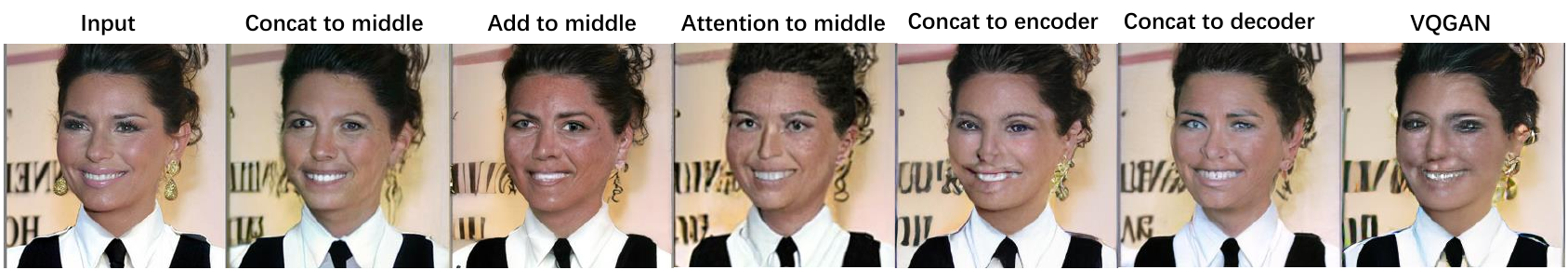}
    \caption{Comparison of different method and position to input embeddings into UNet.}
    \label{fig7}
\end{figure*}

 As show in Figure \ref{fig7}, inputting the image embeddings in the encoding blocks or decoding blocks instead of the middle block, DiVAE reconstrcution nearly exactly the same style image as the VQGAN, mabye model gets too much information from the skip connections when putting into encoding blocks and it's too late to input when putting into decoding blocks. So it is optimal to add embeddings to the middle layer. Based on above, we investigate to add or concat embeddings to middle block, in addition we try to input the embedding by an attention block. As Table \ref{tab3_1} shows, we find that adding and concating don't have much difference, while attention block has poorer performance, maybe it needs more design.Both of adding and concating is effective as show in Figure \ref{fig7}. So the position of inputting embeddings is greatly crucial and concating and adding have little difference.

\section{Conclusion}
In this paper, we proposes a DiVAE with a diffusion decoder  to generate more photorealistic and detailed images to improve the reconstructing stage of multi stage image synthesis. Our model achieves state-of-the-art results on reconstruction of  images comparing with existing approach and samples more detailed images on Text-to-Image tasks. And the potential improvements of the first stage multi stage image synthesis which generates images embedding is expecting to future works. 

\clearpage

\bibliographystyle{plain}
\bibliography{divae.bbl}

\begin{thebibliography}{10}

\bibitem{austin2021structured}
Jacob Austin, Daniel~D Johnson, Jonathan Ho, Daniel Tarlow, and Rianne van~den
  Berg.
\newblock Structured denoising diffusion models in discrete state-spaces.
\newblock {\em Advances in Neural Information Processing Systems},
  34:17981--17993, 2021.

\bibitem{brown2020language}
Tom Brown, Benjamin Mann, Nick Ryder, Melanie Subbiah, Jared~D Kaplan, Prafulla
  Dhariwal, Arvind Neelakantan, Pranav Shyam, Girish Sastry, Amanda Askell,
  et~al.
\newblock Language models are few-shot learners.
\newblock {\em Advances in neural information processing systems},
  33:1877--1901, 2020.

\bibitem{cha2019adversarial}
Miriam Cha, Youngjune~L Gwon, and HT~Kung.
\newblock Adversarial learning of semantic relevance in text to image
  synthesis.
\newblock In {\em Proceedings of the AAAI conference on artificial
  intelligence}, volume~33, pages 3272--3279, 2019.

\bibitem{chen2020generative}
Mark Chen, Alec Radford, Rewon Child, Jeffrey Wu, Heewoo Jun, David Luan, and
  Ilya Sutskever.
\newblock Generative pretraining from pixels.
\newblock In {\em International Conference on Machine Learning}, pages
  1691--1703. PMLR, 2020.

\bibitem{chen2020wavegrad}
Nanxin Chen, Yu~Zhang, Heiga Zen, Ron~J Weiss, Mohammad Norouzi, and William
  Chan.
\newblock Wavegrad: Estimating gradients for waveform generation.
\newblock {\em arXiv preprint arXiv:2009.00713}, 2020.

\bibitem{chen2018neural}
Ricky~TQ Chen, Yulia Rubanova, Jesse Bettencourt, and David~K Duvenaud.
\newblock Neural ordinary differential equations.
\newblock {\em Advances in neural information processing systems}, 31, 2018.

\bibitem{dai2019diagnosing}
Bin Dai and David Wipf.
\newblock Diagnosing and enhancing vae models.
\newblock {\em arXiv preprint arXiv:1903.05789}, 2019.

\bibitem{deng2009imagenet}
Jia Deng, Wei Dong, Richard Socher, Li-Jia Li, Kai Li, and Li~Fei-Fei.
\newblock Imagenet: A large-scale hierarchical image database.
\newblock In {\em 2009 IEEE conference on computer vision and pattern
  recognition}, pages 248--255. Ieee, 2009.

\bibitem{dhariwal2021diffusion}
Prafulla Dhariwal and Alexander Nichol.
\newblock Diffusion models beat gans on image synthesis.
\newblock {\em Advances in Neural Information Processing Systems}, 34, 2021.

\bibitem{ding2021cogview}
Ming Ding, Zhuoyi Yang, Wenyi Hong, Wendi Zheng, Chang Zhou, Da~Yin, Junyang
  Lin, Xu~Zou, Zhou Shao, Hongxia Yang, et~al.
\newblock Cogview: Mastering text-to-image generation via transformers.
\newblock {\em Advances in Neural Information Processing Systems}, 34, 2021.

\bibitem{dinh2014nice}
Laurent Dinh, David Krueger, and Yoshua Bengio.
\newblock Nice: Non-linear independent components estimation.
\newblock {\em arXiv preprint arXiv:1410.8516}, 2014.

\bibitem{esser2021taming}
Patrick Esser, Robin Rombach, and Bjorn Ommer.
\newblock Taming transformers for high-resolution image synthesis.
\newblock In {\em Proceedings of the IEEE/CVF Conference on Computer Vision and
  Pattern Recognition}, pages 12873--12883, 2021.

\bibitem{goodfellow2014generative}
Ian Goodfellow, Jean Pouget-Abadie, Mehdi Mirza, Bing Xu, David Warde-Farley,
  Sherjil Ozair, Aaron Courville, and Yoshua Bengio.
\newblock Generative adversarial nets.
\newblock {\em Advances in neural information processing systems}, 27, 2014.

\bibitem{gu2021vector}
Shuyang Gu, Dong Chen, Jianmin Bao, Fang Wen, Bo~Zhang, Dongdong Chen, Lu~Yuan,
  and Baining Guo.
\newblock Vector quantized diffusion model for text-to-image synthesis.
\newblock {\em arXiv preprint arXiv:2111.14822}, 2021.

\bibitem{heusel2017gans}
Martin Heusel, Hubert Ramsauer, Thomas Unterthiner, Bernhard Nessler, and Sepp
  Hochreiter.
\newblock Gans trained by a two time-scale update rule converge to a local nash
  equilibrium.
\newblock {\em Advances in neural information processing systems}, 30, 2017.

\bibitem{ho2019flow++}
Jonathan Ho, Xi~Chen, Aravind Srinivas, Yan Duan, and Pieter Abbeel.
\newblock Flow++: Improving flow-based generative models with variational
  dequantization and architecture design.
\newblock In {\em International Conference on Machine Learning}, pages
  2722--2730. PMLR, 2019.

\bibitem{ho2020denoising}
Jonathan Ho, Ajay Jain, and Pieter Abbeel.
\newblock Denoising diffusion probabilistic models.
\newblock {\em Advances in Neural Information Processing Systems},
  33:6840--6851, 2020.

\bibitem{ho2022cascaded}
Jonathan Ho, Chitwan Saharia, William Chan, David~J Fleet, Mohammad Norouzi,
  and Tim Salimans.
\newblock Cascaded diffusion models for high fidelity image generation.
\newblock {\em Journal of Machine Learning Research}, 23(47):1--33, 2022.

\bibitem{ho2021classifier}
Jonathan Ho and Tim Salimans.
\newblock Classifier-free diffusion guidance.
\newblock In {\em NeurIPS 2021 Workshop on Deep Generative Models and
  Downstream Applications}, 2021.

\bibitem{kingma2013auto}
Diederik~P Kingma and Max Welling.
\newblock Auto-encoding variational bayes.
\newblock {\em arXiv preprint arXiv:1312.6114}, 2013.

\bibitem{kingma2018glow}
Durk~P Kingma and Prafulla Dhariwal.
\newblock Glow: Generative flow with invertible 1x1 convolutions.
\newblock {\em Advances in neural information processing systems}, 31, 2018.

\bibitem{kong2020diffwave}
Zhifeng Kong, Wei Ping, Jiaji Huang, Kexin Zhao, and Bryan Catanzaro.
\newblock Diffwave: A versatile diffusion model for audio synthesis.
\newblock {\em arXiv preprint arXiv:2009.09761}, 2020.

\bibitem{lao2019dual}
Qicheng Lao, Mohammad Havaei, Ahmad Pesaranghader, Francis Dutil, Lisa~Di
  Jorio, and Thomas Fevens.
\newblock Dual adversarial inference for text-to-image synthesis.
\newblock In {\em Proceedings of the IEEE/CVF International Conference on
  Computer Vision}, pages 7567--7576, 2019.

\bibitem{li2019controllable}
Bowen Li, Xiaojuan Qi, Thomas Lukasiewicz, and Philip Torr.
\newblock Controllable text-to-image generation.
\newblock {\em Advances in Neural Information Processing Systems}, 32, 2019.

\bibitem{liang2020cpgan}
Jiadong Liang, Wenjie Pei, and Feng Lu.
\newblock Cpgan: Content-parsing generative adversarial networks for
  text-to-image synthesis.
\newblock In {\em European Conference on Computer Vision}, pages 491--508.
  Springer, 2020.

\bibitem{lin2021m6}
Junyang Lin, Rui Men, An~Yang, Chang Zhou, Ming Ding, Yichang Zhang, Peng Wang,
  Ang Wang, Le~Jiang, Xianyan Jia, et~al.
\newblock M6: A chinese multimodal pretrainer.
\newblock {\em arXiv preprint arXiv:2103.00823}, 2021.

\bibitem{lin2014microsoft}
Tsung-Yi Lin, Michael Maire, Serge Belongie, James Hays, Pietro Perona, Deva
  Ramanan, Piotr Doll{\'a}r, and C~Lawrence Zitnick.
\newblock Microsoft coco: Common objects in context.
\newblock In {\em European conference on computer vision}, pages 740--755.
  Springer, 2014.

\bibitem{nichol2021glide}
Alex Nichol, Prafulla Dhariwal, Aditya Ramesh, Pranav Shyam, Pamela Mishkin,
  Bob McGrew, Ilya Sutskever, and Mark Chen.
\newblock Glide: Towards photorealistic image generation and editing with
  text-guided diffusion models.
\newblock {\em arXiv preprint arXiv:2112.10741}, 2021.

\bibitem{nichol2021improved}
Alexander~Quinn Nichol and Prafulla Dhariwal.
\newblock Improved denoising diffusion probabilistic models.
\newblock In {\em International Conference on Machine Learning}, pages
  8162--8171. PMLR, 2021.

\bibitem{ramesh2022hierarchical}
Aditya Ramesh, Prafulla Dhariwal, Alex Nichol, Casey Chu, and Mark Chen.
\newblock Hierarchical text-conditional image generation with clip latents.
\newblock {\em arXiv preprint arXiv:2204.06125}, 2022.

\bibitem{ramesh2021zero}
Aditya Ramesh, Mikhail Pavlov, Gabriel Goh, Scott Gray, Chelsea Voss, Alec
  Radford, Mark Chen, and Ilya Sutskever.
\newblock Zero-shot text-to-image generation.
\newblock In {\em International Conference on Machine Learning}, pages
  8821--8831. PMLR, 2021.

\bibitem{rezende2014stochastic}
Danilo~Jimenez Rezende, Shakir Mohamed, and Daan Wierstra.
\newblock Stochastic backpropagation and approximate inference in deep
  generative models.
\newblock In {\em International conference on machine learning}, pages
  1278--1286. PMLR, 2014.

\bibitem{rombach2021high}
Robin Rombach, Andreas Blattmann, Dominik Lorenz, Patrick Esser, and Bj{\"o}rn
  Ommer.
\newblock High-resolution image synthesis with latent diffusion models.
\newblock {\em arXiv preprint arXiv:2112.10752}, 2021.

\bibitem{rombach2020network}
Robin Rombach, Patrick Esser, and Bjorn Ommer.
\newblock Network-to-network translation with conditional invertible neural
  networks.
\newblock {\em Advances in Neural Information Processing Systems},
  33:2784--2797, 2020.

\bibitem{ronneberger2015u}
Olaf Ronneberger, Philipp Fischer, and Thomas Brox.
\newblock U-net: Convolutional networks for biomedical image segmentation.
\newblock In {\em International Conference on Medical image computing and
  computer-assisted intervention}, pages 234--241. Springer, 2015.

\bibitem{salimans2017pixelcnn++}
Tim Salimans, Andrej Karpathy, Xi~Chen, and Diederik~P Kingma.
\newblock Pixelcnn++: Improving the pixelcnn with discretized logistic mixture
  likelihood and other modifications.
\newblock {\em arXiv preprint arXiv:1701.05517}, 2017.

\bibitem{sohl2015deep}
Jascha Sohl-Dickstein, Eric Weiss, Niru Maheswaranathan, and Surya Ganguli.
\newblock Deep unsupervised learning using nonequilibrium thermodynamics.
\newblock In {\em International Conference on Machine Learning}, pages
  2256--2265. PMLR, 2015.

\bibitem{song2020denoising}
Jiaming Song, Chenlin Meng, and Stefano Ermon.
\newblock Denoising diffusion implicit models.
\newblock {\em arXiv preprint arXiv:2010.02502}, 2020.

\bibitem{tan2019semantics}
Hongchen Tan, Xiuping Liu, Xin Li, Yi~Zhang, and Baocai Yin.
\newblock Semantics-enhanced adversarial nets for text-to-image synthesis.
\newblock In {\em Proceedings of the IEEE/CVF International Conference on
  Computer Vision}, pages 10501--10510, 2019.

\bibitem{tao2020df}
Ming Tao, Hao Tang, Songsong Wu, Nicu Sebe, Xiao-Yuan Jing, Fei Wu, and Bingkun
  Bao.
\newblock Df-gan: Deep fusion generative adversarial networks for text-to-image
  synthesis.
\newblock {\em arXiv preprint arXiv:2008.05865}, 2020.

\bibitem{van2016conditional}
Aaron Van~den Oord, Nal Kalchbrenner, Lasse Espeholt, Oriol Vinyals, Alex
  Graves, et~al.
\newblock Conditional image generation with pixelcnn decoders.
\newblock {\em Advances in neural information processing systems}, 29, 2016.

\bibitem{van2017neural}
Aaron Van Den~Oord, Oriol Vinyals, et~al.
\newblock Neural discrete representation learning.
\newblock {\em Advances in neural information processing systems}, 30, 2017.

\bibitem{van2016pixel}
Aaron Van~Oord, Nal Kalchbrenner, and Koray Kavukcuoglu.
\newblock Pixel recurrent neural networks.
\newblock In {\em International conference on machine learning}, pages
  1747--1756. PMLR, 2016.

\bibitem{wu2021godiva}
Chenfei Wu, Lun Huang, Qianxi Zhang, Binyang Li, Lei Ji, Fan Yang, Guillermo
  Sapiro, and Nan Duan.
\newblock Godiva: Generating open-domain videos from natural descriptions.
\newblock {\em arXiv preprint arXiv:2104.14806}, 2021.

\bibitem{wu2021n}
Chenfei Wu, Jian Liang, Lei Ji, Fan Yang, Yuejian Fang, Daxin Jiang, and Nan
  Duan.
\newblock N$\backslash$" uwa: Visual synthesis pre-training for neural visual
  world creation.
\newblock {\em arXiv preprint arXiv:2111.12417}, 2021.

\bibitem{xu2018attngan}
Tao Xu, Pengchuan Zhang, Qiuyuan Huang, Han Zhang, Zhe Gan, Xiaolei Huang, and
  Xiaodong He.
\newblock Attngan: Fine-grained text to image generation with attentional
  generative adversarial networks.
\newblock In {\em Proceedings of the IEEE conference on computer vision and
  pattern recognition}, pages 1316--1324, 2018.

\bibitem{yu2021vector}
Jiahui Yu, Xin Li, Jing~Yu Koh, Han Zhang, Ruoming Pang, James Qin, Alexander
  Ku, Yuanzhong Xu, Jason Baldridge, and Yonghui Wu.
\newblock Vector-quantized image modeling with improved vqgan.
\newblock {\em arXiv preprint arXiv:2110.04627}, 2021.

\bibitem{zhang2021cross}
Han Zhang, Jing~Yu Koh, Jason Baldridge, Honglak Lee, and Yinfei Yang.
\newblock Cross-modal contrastive learning for text-to-image generation.
\newblock In {\em Proceedings of the IEEE/CVF Conference on Computer Vision and
  Pattern Recognition}, pages 833--842, 2021.

\bibitem{zhu2019dm}
Minfeng Zhu, Pingbo Pan, Wei Chen, and Yi~Yang.
\newblock Dm-gan: Dynamic memory generative adversarial networks for
  text-to-image synthesis.
\newblock In {\em Proceedings of the IEEE/CVF Conference on Computer Vision and
  Pattern Recognition}, pages 5802--5810, 2019.

\end{thebibliography}


\clearpage
\appendix


\section{Training}
We train the model on ImageNet datasets on 32 A100 GPUs with a training batch size of 256 for total 10k training steps, AdamW optimizer is used with the learning rate linearly warming up to a peak value of 1e-4 over 1000 steps.

\section{Limitations}
We note that our model still has more time consumption than un-diffusion model, so we also experiment DDIM sample approach with 25 steps. From the comparison on reconstructing tasks, we find that the FID of DiVAE with DDIM 25steps can't be better than DDPM 1000 steps. However, in comparison of visual effect, DiVAE is more detailed than VQGAN no matter in DDPM or DDIM sample mode. The time consuming problem of diffusion model need to be further solved. In addition, the size of generative image is stationary, which is determined by training data. A trained model can't accept and output various size images as VQ-GAN. The variability of synthesis image's size is worth exploring in further works.

\begin{table*}[h]
\setlength{\tabcolsep}{0.16in}
\label{table3}
\begin{center}
\begin{tabular}{lcccccr}
\toprule
Model & Dataset & $D \rightarrow R$ & Rate & dim $Z$ &FID \\
\midrule
DALL-E dVAE & Web data & $32^2 \rightarrow 256^2$ & f8 & 8192 &32.0 \\
VQGAN & ImageNet & $16^2 \rightarrow 256^2$ & f16 & 1024&7.94 \\
VQGAN & ImageNet & $16^2 \rightarrow 256^2$ & f16 & 16384&4.98 \\
VQGAN & ImageNet & $32^2 \rightarrow 256^2$ & f8 & 8192&1.49\\
\midrule
DiVAE & ImageNet & $16^2 \rightarrow 256^2$ & f16 & 16384& \textbf{4.07}\\
DiVAE & ImageNet & $32^2 \rightarrow 256^2$ & f8 & 8192& \textbf{1.24} \\
DiVAE(ddim25) & ImageNet & $16^2 \rightarrow 256^2$ & f16 & 16384& 7.07\\
DiVAE(ddim25) & ImageNet & $32^2 \rightarrow 256^2$ & f8 & 8192& 3.14 \\
\bottomrule
\end{tabular}
\end{center}
\vskip -0.05in
\caption{FID between reconstructed validation split and original validation with 50000 images split on ImageNet.}
\label{restruct}
\end{table*}

\begin{figure*}[h]
    \centering
    \includegraphics[width=4in]{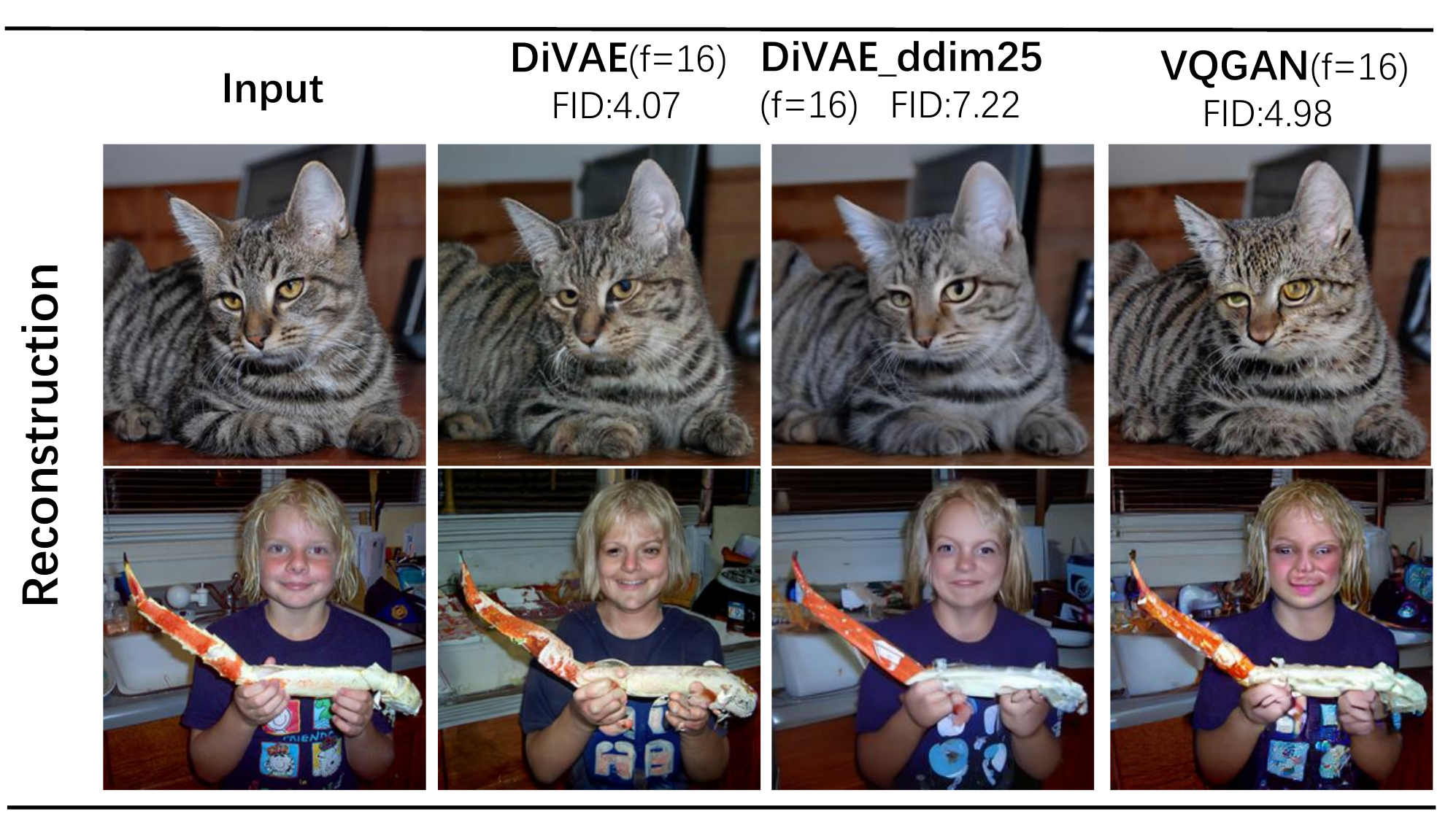}
    \caption{Comparison with DDIM sample approach .}
    \label{fig11}
\end{figure*}

\section{Additional Results}
\begin{figure*}[h]
    \centering
    \includegraphics[width=5.5in]{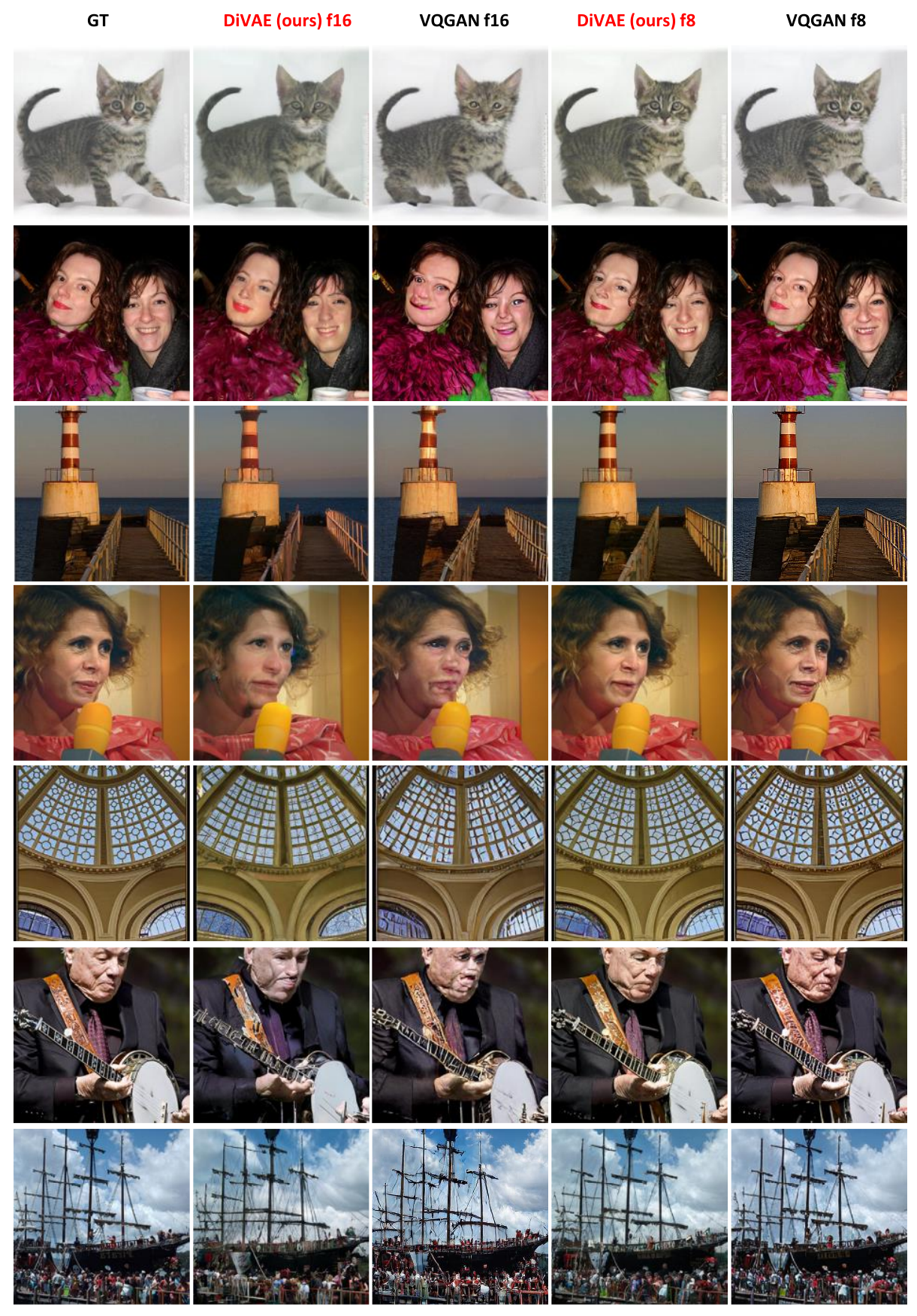}
    \caption{Comparison with VQGAN in $32^2 \rightarrow 256^2$ and $16^2 \rightarrow 256^2$ reconstruction with same codebook dimension $Z$.}
    \label{fig8}
\end{figure*}

\begin{figure*}[h]
    \centering
    \includegraphics[width=5.5in]{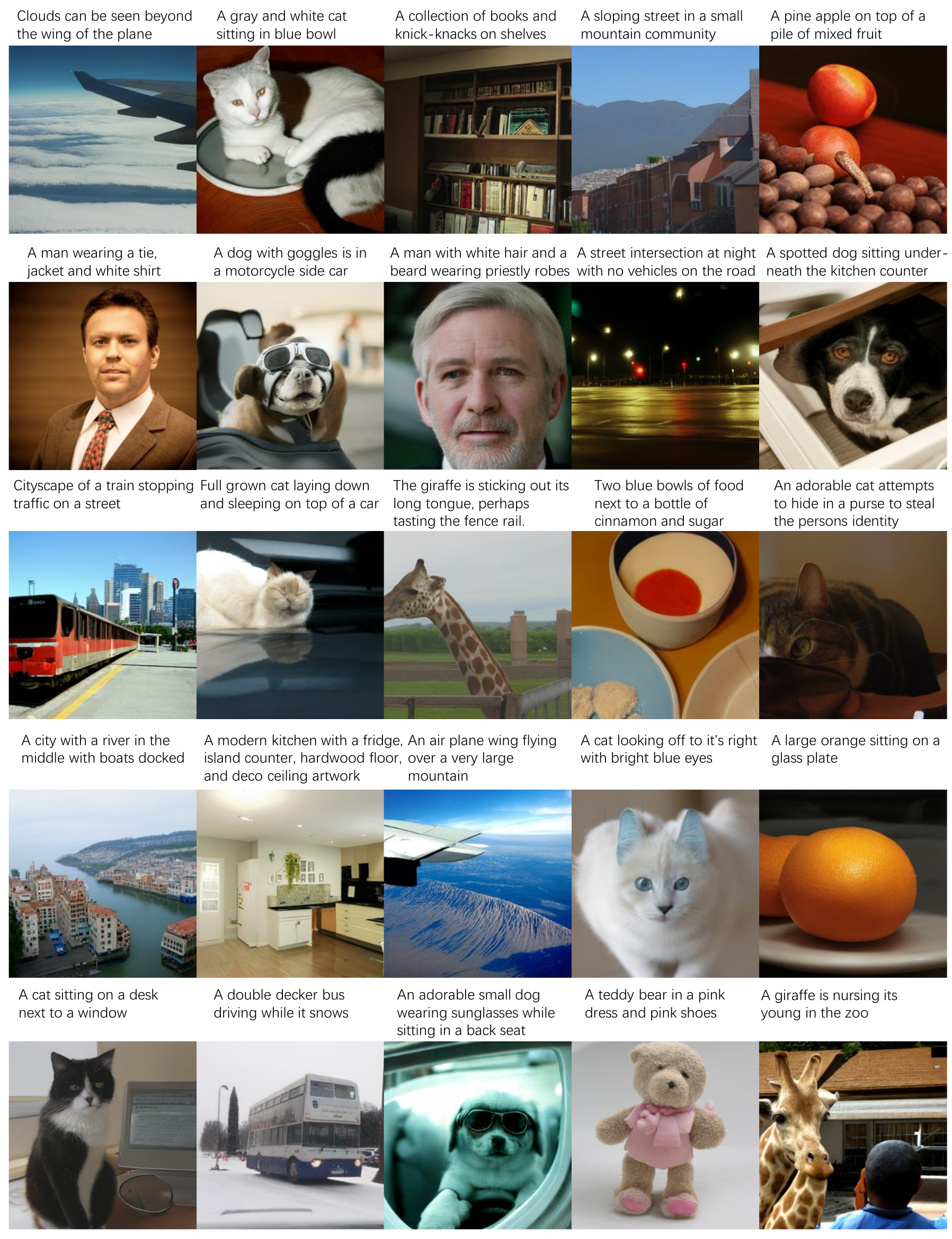}
    \caption{Text-to-Image 256×256 samples.}
    \label{fig9}
\end{figure*}

In this part, we provide more samples of comparison on reconstruction in Figure \ref{fig8} and Text-to-Image task in Figure \ref{fig9}.

\end{document}